\documentclass{article}
\usepackage{spconf,amsmath,graphicx,setspace}
\usepackage{amssymb}
\usepackage{algorithm}
\usepackage{algorithmic}
\usepackage{textcomp}
\usepackage{subfigure}
\usepackage{tipa}
\usepackage{cite} 
\usepackage{url}
\usepackage{float}

\makeatletter 
\renewcommand{\@thesubfigure}{\hskip\subfiglabelskip}
\makeatother
\title{TRIBYOL: TRIPLET BYOL FOR SELF-SUPERVISED REPRESENTATION LEARNING}
\name{Guang Li $^{\dagger}$ \qquad Ren Togo $^{\dagger\dagger}$ \qquad Takahiro Ogawa $^{\dagger\dagger\dagger}$ \qquad Miki Haseyama$^{\dagger\dagger\dagger}$ \thanks{
This work was partly supported by AMED Grant Number JP21zf0127004. This  study  was  conducted  on  the  Data  Science  Computing System of Education and Research Center for Mathematical and Data Science, Hokkaido University.}}
\address{$^{\dagger}$ Graduate School of Information Science and Technology,
    Hokkaido University, Japan \\
    $^{\dagger\dagger}$ Education and Research Center for Mathematical and Data Science,
 	Hokkaido University, Japan \\
    $^{\dagger\dagger\dagger}$ Faculty of Information Science and Technology, 
    Hokkaido University, Japan \\
 	E-mail: \{guang, togo, ogawa\}@lmd.ist.hokudai.ac.jp, miki@ist.hokudai.ac.jp}
\begin{document}
\ninept
\maketitle
%
\begin{abstract}
\end{abstract}
This paper proposes a novel self-supervised learning method for learning better representations with small batch sizes.
Many self-supervised learning methods based on certain forms of the siamese network have emerged and received significant attention.
However, these methods need to use large batch sizes to learn good representations and require heavy computational resources.
We present a new triplet network combined with a triple-view loss to improve the performance of self-supervised representation learning with small batch sizes.
Experimental results show that our method can drastically outperform state-of-the-art self-supervised learning methods on several datasets in small-batch cases.
Our method provides a feasible solution for self-supervised learning with real-world high-resolution images that uses small batch sizes.
\par
\section{Introduction}
%
Deep supervised learning has shown outstanding performance in many areas, especially on various computer vision tasks such as image classification, 
object detection and semantic segmentation~\cite{lecun2015deep}.  
However, supervised learning methods have over-reliance on manually designed labels and suffer from generalization problems, and hence are meeting their bottlenecks~\cite{liu2020self}.
As an alternative, self-supervised learning is a learning framework that conforms to human cognition, which can learn information from the data itself without the need for manually designed labels~\cite{jing2020self}.
Self-supervised learning has shown performance comparable to supervised learning methods on multiple tasks and has received widespread attention.
\par
Early self-supervised learning methods often learn representations via a pretext task that applies a transformation to the input image and requires the learner to predict the properties of transformation ($e.g.$, rotation~\cite{gidaris2018unsupervised} and jigsaw~\cite{noroozi2016unsupervised}) from the transformed image~\cite{jing2020self}.
Although such transformations are beneficial for predicting 3D correlations, it is undesirable for most semantic recognition tasks.
PIRL~\cite{misra2020self} proposes to learn invariant representations rather than covariant ones.
The method constructs image representations similar to the representations of transformed versions of the same image and different from the representations of other images.
By combining the jigsaw or rotation pretext task with PIRL, the performance surpassed supervised learning in some computer vision tasks.
\par
Recently, self-supervised learning methods based on the siamese network~\cite{Bromley1993signature} achieved high representation learning performance on natural images.
These methods define the inputs as two augmented views from one image, then input to the siamese network and maximize the similarity between the representations of views~\cite{tian2020understanding}. 
When the batch size decreases, these methods face accuracy degradation problems due to the reduced mutual information and transformation-invariant representation~\cite{li2021cross}.
Therefore, these methods typically need to increase the number of samples in a batch and need heavy computational resources ($e.g.$, at least 4 GPUs or 32 TPU cores) for learning better representations from images~\cite{tian2021understanding}.
For example, the state-of-the-art methods,
SimCLR~\cite{chen2020simple}, SimSiam~\cite{chen2021exploring}, and BYOL~\cite{grill2020bootstrap} all use batch sizes over 128. 
Requiring such a huge batch size is expensive and impractical in the real-world ($e.g.$, high-resolution medical images~\cite{li2021self, li2021triplet, li2022self2} and remote sensing images~\cite{tao2020remote}).
Hence, it is crucial to explore self-supervised learning in small-batch cases.
\par
In this paper, we propose a novel self-supervised learning method called TriBYOL for learning better representations with small batch sizes.
Considering the appropriate addition of augmented views can increase mutual information and encourage a more transformation-invariant representation in small-batch cases~\cite{tian2020contrastive}, we present a new triplet network combined with a triple-view loss based on the state-of-the-art method BYOL~\cite{grill2020bootstrap}.
The conventional triplet network~\cite{hoffer2015deep} is a variant of the siamese network and typically has three inputs, including anchor, positive and negative samples.
Different from the conventional triplet network, our method only has three augmented views from one image.
Experimental results show that our method can drastically outperform the state-of-the-art self-supervised learning methods on several datasets in small-batch cases.
\par
Our main contributions can be summarized as follows.
\begin{itemize}
    \item We propose a novel self-supervised learning method called TriBYOL for learning better representations with small batch sizes.
    \item We confirm that our method can drastically outperform state-of-the-art self-supervised learning methods on several datasets in small-batch cases.
\end{itemize}
\section{Methodology}
\begin{figure*}[t]
        \centering
        \includegraphics[width=15.0cm]{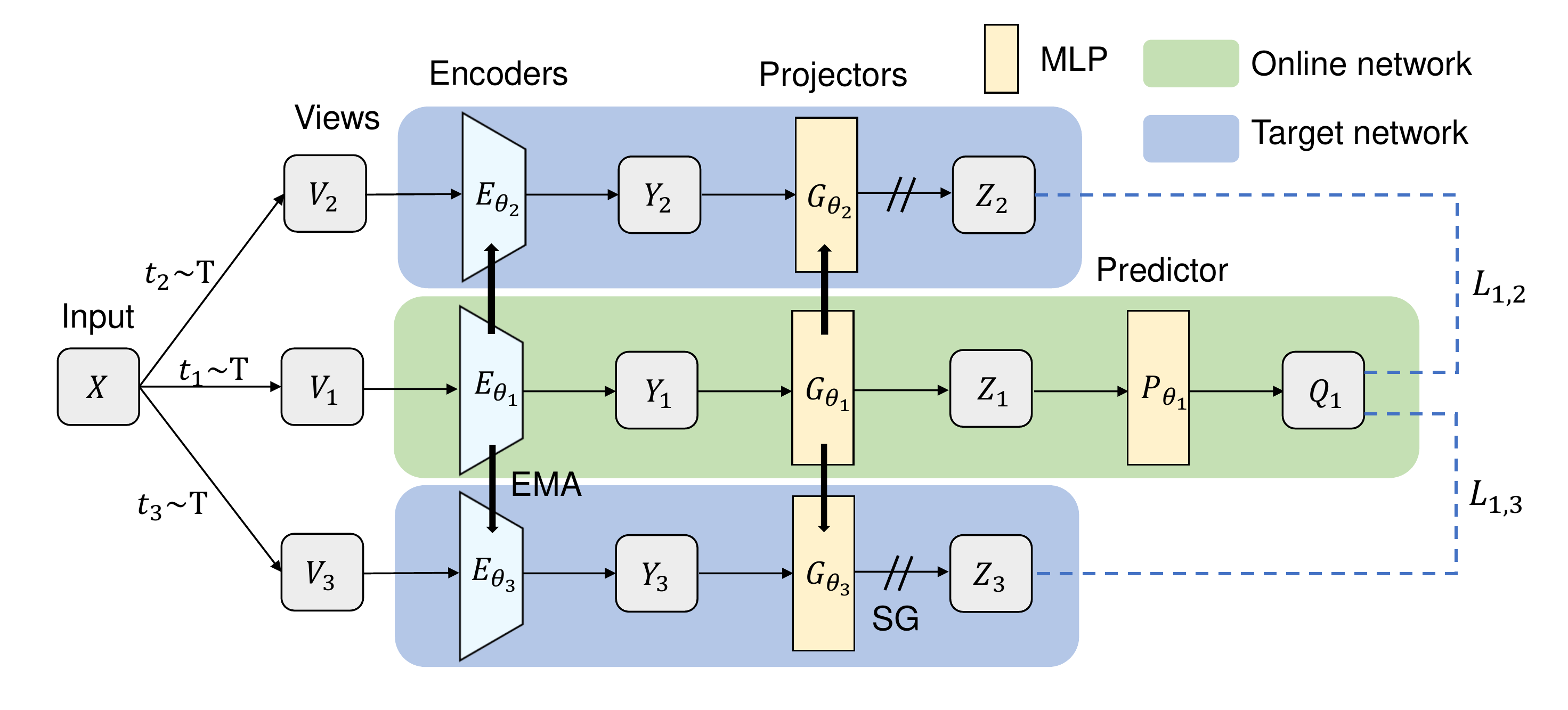}
        \caption{Overview of the proposed method. Our method minimizes a triple-view loss between representations of three views from the triplet network. The weights of the target networks ($\theta_{2}$ and $\theta_{3}$) are an exponential moving average (EMA) of the weights of the online network ($\theta_{1}$). MLP denotes multilayer perceptron. SG denotes stop-gradient.}
        \label{fig1}
\end{figure*}
\subsection{Description of TriBYOL}
The overview of our method is shown in Fig.~\ref{fig1}.
Motivated by the fact that appropriate addition of augmented views can increase mutual information and encourage a more transformation-invariant representation in small-batch cases~\cite{tian2020contrastive}, different from BYOL~\cite{grill2020bootstrap} which uses the siamese network, we propose the triplet network combined with a triple-view loss for learning better representations with small batch sizes.
Our method uses a triplet network to learn discriminative representations from images.
The triplet network contains one online network and two target networks with the same structure, where the weights of the target networks are an exponential moving average (EMA) of the weights of the online network~\cite{antti2017mean}.
The conventional triplet network is a variant of the siamese network and typically has three inputs, including anchor, positive and negative samples.
Different from the conventional triplet network, our method only has three augmented views from one image.
\par
Encoder $E_{\theta_{1}}$ and predictor $G_{\theta_{1}}$ belong to the online network.
Encoders $E_{\theta_{2}}$ and $E_{\theta_{3}}$ belong to the target networks.
Given an input image $X$ from a dataset $D$ without label information, three transformations $t_{1}$, $t_{2}$, and $t_{3}$ are randomly sampled from a distribution $T$ to generate three views $V_{1} = t_{1}(X)$, $V_{2} = t_{2}(X)$, and $V_{3} = t_{3}(X)$.
Similar to BYOL, our method exchanges views for learning better representations.
$Y_{1}$, $Y_{2}$, and $Y_{3}$ are representations processed by the encoders.
$Z_{1}$, $Z_{2}$, and $Z_{3}$ are representations processed by the projector.
$Q_{1}$ is representations processed by the predictor.
Note that we only use the predictor in the online network to make the triplet network asymmetric, which can prevent learning from collapsing, thereby improving the representation learning performance~\cite{grill2020bootstrap}.
Finally, we define the loss $L$ to compare the normalized representations from different views of the same image.
The triple-view loss $L$ comparing representations of $V_{1}$, $V_{2}$, and $V_{3}$ is defined as follows:
\begin{equation}
\label{equ1}
\begin{split}
L_{i,j} 
& = 
|| \hat{Q}_{i} - \hat{Z}_{j} ||_{2}^{2},
\\ & = 2 - 2 \cdot \frac{\left \langle Q_{i},Z_{j} \right \rangle}{ || Q_{i} ||_{2} \cdot || Z_{j} ||_{2}},
\end{split}
\end{equation}
\begin{equation}
\label{equ2}
L = L_{1,2} + L_{1,3},
\end{equation}
where $\hat{Q}_{i} = Q_{i}/ || Q_{i} ||_{2}$ and $\hat{Z}_{i} = Z_{i}/ || Z_{i} ||_{2}$ denote the normalized representations of $V_{i}$ ($i={1,2,3}$). 
The consistency among views from the triplet network helps to learn discriminative representations from images.
The weights of the online network ($\theta_{1}$) are updated as follows:
\begin{equation}
\label{equ3}
\theta_{1} \leftarrow \mathrm{Opt}(\theta_{1}, \nabla_{\theta_{1}}{L}, \alpha),
\end{equation}
where $\mathrm{Opt}$ and $\alpha$ denote the optimizer and the learning rate, respectively.
Since the stop-gradient (SG) is critical for preventing the collapse of self-supervised learning~\cite{chen2021exploring}, the target networks ($E_{\theta_{2}}$ and $E_{\theta_{3}}$) are not updated using backpropagation. 
The weights of target networks ($\theta_{2}$ and $\theta_{3}$) are an exponential moving average of the weights of $\theta_{1}$ and are updated as follows:
\begin{equation}
\label{equ4}
\theta_{2} \leftarrow \tau\theta_{2} + (1-\tau)\theta_{1},
\end{equation}
\begin{equation}
\label{equ5}
\theta_{3} \leftarrow \tau\theta_{3} + (1-\tau)\theta_{1},
\end{equation}
where $\tau$ denotes a momentum coefficient, and we alternately update the weights $\theta_{2}$ and $\theta_{3}$ after every iteration.
After the self-supervised learning, we can use the trained encoder ($E_{\theta_{1}}$) for downstream tasks such as linear evaluation and fine-tuning.
The self-supervised learning process of our method is summarized in Algorithm~\ref{alg1}.
\begin{algorithm}[t]
    \caption{TriBYOL}    
    \label{alg1}
    \begin{algorithmic}[1]
    \REQUIRE 
    $D$: dataset; 
    $B$: batch size;
    $T$: distribution;
    $E$: training epochs
    \ENSURE
    $E_{\theta_{1}}$: the trained encoder 
    \\
    \FOR{each training epoch $t = 1$ to $E$}
    \STATE
    Load a minibatch $X$ from $D$ with $B$ samples  
    \STATE
    Sample three transformations from $T$ as $t_{1}$, $t_{2}$, and $t_{3}$
    \STATE
    Generate views with random augmentation: \\
    $V_{1} = t_{1}(X)$, $V_{2} = t_{2}(X)$, and $V_{3} = t_{3}(X)$
    \STATE
    Calculate the triple-view loss $L$ with Eqs. (1) and (2)
    \STATE
    Update the weights of the online network ($\theta_{1}$) with Eq. (3)
    \STATE
    Update the weights of the target networks ($\theta_{2}$ and $\theta_{3}$) with Eqs. (4) and (5)
    \ENDFOR
    \end{algorithmic}
\end{algorithm}
\begin{table*}[t]
    \centering
    \caption{Linear evaluation results (\%) with different batch sizes. ``b32'', ``b64'', and ``b128'' denote the batch size of 32, 64, and 128, respectively.}
    \label{tab1}
    \begin{tabular}{c|ccc|ccc|ccc}
    \hline
    & & CIFAR-10 & & & CIFAR-100 & & & STL-10\\\hline
    Method & b32 & b64 & b128 & b32 & b64 & b128 & b32 & b64 & b128 \\\hline
    TriBYOL 
    & \bfseries{79.09} & \bfseries{85.35} & \bfseries{87.31}
    & \bfseries{49.07} & \bfseries{59.90} & \bfseries{63.05}
    & \bfseries{75.41} & \bfseries{83.16} & \bfseries{88.19} \\
    Cross
    & 76.01 & 82.06 & 83.50
    & 48.04 & 54.65 & 58.79 
    & 69.66 & 78.38 & 83.79 \\
    BYOL
    & 68.67 & 81.47 & 83.79
    & 41.21 & 49.68 & 58.34
    & 49.60 & 80.09 & 84.88 \\
    SimSiam
    & 58.42 & 71.25 & 75.58
    & 1.00 & 37.06 & 49.21
    & 10.00 & 65.20 & 71.78 \\\hline
    PIRL-rotation
    & - & - & 55.78
    & - & - & 31.55
    & - & - & 50.26 \\
    PIRL-jigsaw
    & - & - & 49.94
    & - & - & 27.36
    & - & - & 48.55 \\
    SimCLR
    & - & - & 52.58
    & - & - & 21.26
    & - & - & 44.50 \\\hline\hline
    ImageNet
    &  & 82.37 & 
    &  & 60.72 & 
    &  & 91.76 & \\
    \hline
    \end{tabular}
\end{table*}
\subsection{Self-supervised learning in small-batch cases}
In this subsection, we propose a new problem setting and discuss its feasibility.
Self-supervised learning methods based on the siamese network have achieved high representation learning performance on natural images.
However, these methods typically need to increase the number of samples in a batch for learning better representations from images.
Requiring such a huge batch size is expensive and impractical in the real world.
For example, in practical applications, high-resolution medical images and remote sensing images are difficult to perform self-supervised learning with large batch sizes due to the limitation of GPU memory.
To solve the above problem, we propose the triplet network combined with a triple-view loss based on the state-of-the-art method BYOL~\cite{grill2020bootstrap}.
The appropriate addition of augmented views can increase mutual information and encourage a more transformation-invariant representation in small-batch cases~\cite{tian2020contrastive}.
Our method can provide a feasible solution for subsequent practical applications ($e.g.$, high-resolution medical images and remote sensing images) of self-supervised learning in small-batch cases.
\subsection{Implementation details}
$\mathbf{Image\,\,Augmentations}$
Three random views are generated by a combination of standard data augmentation methods, including cropping, resizing, flipping, color jittering, grayscaling, and Gaussian blurring based on BYOL~\cite{grill2020bootstrap} data augmentation pipeline.
The following operations are performed sequentially to produce each view.
\begin{itemize}
    \item Random cropping with an area uniformly sampled with a scale between 0.2 to 1.0, followed by resizing to a resolution of 96 $\times$ 96 pixels.
    \item Random horizontal flipping with an applying probability of 0.5.
    \item Color jittering of brightness, contrast, saturation, and hue with the strength of 0.8, 0.8, 0.8, and 0.2 with an applying probability of 0.8.
    \item Grayscaling with an applying probability of 0.2.
    \item Gaussian blurring with kernel size 9 and std between 0.1 to 2.0.
\end{itemize}
$\mathbf{Architecture}$
The encoders ($E_{\theta_{1}}$, $E_{\theta_{2}}$, and $E_{\theta_{3}}$) in our method are ResNet50~\cite{he2016deep} backbone.
The projectors ($G_{\theta_{1}}$, $G_{\theta_{2}}$, and $G_{\theta_{3}}$) are multilayer perceptron (MLP) whose architecture contains a linear layer with an output size of 512, a batch normalization layer, a ReLU activation function, and a linear layer with an output size of 128.
The predictor ($G_{\theta_{1}}$) is also an MLP with the same architecture.
\\\\
$\mathbf{Optimization}$
The optimizer ($\mathrm{Opt}$) used in our method is an SGD optimizer, whose learning rate ($\alpha$), momentum, and weight decay are set to 0.03, 0.9, 0.0004, respectively.
We search for learning rate, momentum, and weight decay based on 40-epoch results and then apply it for longer learning such as 80 epochs or 200 epochs.
We use different small batch sizes such as 32, 64, and 128, which are friendly to typical single GPU implementations.
\section{Experiments}
\begin{table*}[t]
    \centering
    \caption{Fine-tuning results (\%) with different numbers of labels. ``1\%'', ``10\%'', and ``100\%'' denote the percentage of used labels.}
    \label{tab2}
    \begin{tabular}{c|ccc|ccc|ccc}
    \hline
    & & CIFAR-10 & & & CIFAR-100 & & & STL-10 &\\\hline
    Method & 1\% & 10\% & 100\% & 1\% & 10\% & 100\% & 1\% & 10\% & 100\% \\\hline
    TriBYOL 
    & \bfseries{56.60} & \bfseries{71.73} & \bfseries{87.07}
    & \bfseries{9.50} & \bfseries{23.57} & \bfseries{58.92} 
    & \bfseries{56.66} & \bfseries{67.72} & \bfseries{97.34} \\ 
    Cross
    & 50.88 & 67.34 & 86.03
    & 6.81 & 20.96 & 57.23
    & 42.80 & 59.22 & 93.28 \\
    BYOL
    & 56.28 & 70.94 & 86.87
    & 9.38 & 22.51 & 58.17
    & 53.96 & 65.98 & 97.26 \\
    SimSiam
    & 43.16 & 63.08 & 84.76
    & 4.86 & 14.76 & 54.70
    & 40.38 & 49.96 & 88.38 \\
    From Scratch
    & 32.29 & 57.24 & 83.87
    & 5.95 & 17.47 & 56.70 
    & 20.38 & 39.10 & 83.96 \\\hline\hline
    ImageNet 
    & 69.99 & 84.27 & 91.29
    & 27.48 & 52.41 & 70.80
    & 81.24 & 86.34 & 98.00 \\
    \hline
    \end{tabular}
\end{table*}
\begin{table*}[t]
    \centering
    \caption{Transfer learning results (\%) on different datasets.}
    \label{tab3}
    \begin{tabular}{c|c|c|c|c|c|c|c|c}
    \hline
    & MNIST & FashionMNIST & KMNIST & USPS & SVHN & CIFAR-10 & CIFAR-100 & STL-10 \\\hline
    TriBYOL
    & \bfseries{98.74} & \bfseries{91.76} & \bfseries{92.00} & \bfseries{96.61} & \bfseries{75.23} & \bfseries{80.09} & \bfseries{55.88} & \bfseries{79.11} \\
    Cross
    & 98.54 & 91.28 & 90.33 & 96.21 & 71.29 & 77.55 & 51.53 & 76.26 \\
    BYOL
    & 98.41 & 90.77 & 89.88 & 96.06 & 68.75 & 75.31 & 48.51 & 74.04 \\
    SimSiam
    & 97.58 & 89.25 & 83.31 & 94.02 & 58.70 & 64.51 & 35.63 & 63.44 \\\hline\hline
    ImageNet
    & 98.58 & 90.85 & 90.77 & 96.56 & 77.34 & 82.37 & 60.72 & 91.76 \\
    \hline
    \end{tabular}
\end{table*}
In this section, we conduct experiments to verify the effectiveness of TriBYOL.
First, we show the experimental settings in section 3.1.
Next, we show the linear evaluation results, fine-tuning results, and transfer learning results in sections 3.2, 3.3, and 3.4, respectively.
All of our experiments were conducted by using the PyTorch framework with an NVIDIA Tesla P100 GPU that has 16G memory.
\subsection{Experimental settings}
In the experiments, we used 9 benchmark datasets to verify the effectiveness of TriBYOL including MNIST~\cite{lecun2010mnist}, FashionMNIST~\cite{xiao2017fashion},
KMNIST~\cite{clanuwat2018deep},
USPS~\cite{hull1994database},
SVHN~\cite{netzer2011reading}, CIFAR-10~\cite{krizhevsky2009learning}, CIFAR-100~\cite{krizhevsky2009learning}, STL-10~\cite{coates2011analysis}, and
Tiny ImageNet~\cite{le2015tiny}. 
We used training images without label information for self-supervised learning on well-known MNIST, FashionMNIST, KMNIST, USPS, SVHN, CIFAR-10, and CIFAR100.
The STL-10 dataset contains 100,000 unlabeled images and 13,000 labeled images from 10 classes, among which 5,000 images are for training while the remaining 8,000 images are for testing.
For STL-10, we used 100,000 unlabeled images and 5,000 training images without label information for self-supervised learning.
The training set of the Tiny ImageNet dataset contains 100,000 images of 200 classes (500 images for each class).
We only used the training set of Tiny ImageNet without label information for self-supervised learning and transfer learning performance test in subsection 3.4.
\par
To verify the effectiveness of TriBYOL, we use the following methods as comparative methods.
Cross~\cite{li2021cross}, BYOL~\cite{grill2020bootstrap}, and SimSiam~\cite{chen2021exploring} are state-of-the-art self-supervised learning methods without negative sample pairs.
PIRL-rotation~\cite{misra2020self}, PIRL-jigsaw~\cite{misra2020self}, and SimCLR~\cite{chen2020simple} are state-of-the-art self-supervised learning methods with negative sample pairs.
ImageNet and From Scratch denote pretraining on ImageNet~\cite{deng2009imagenet} and training from scratch ($i.e.$, random initial weights), respectively.
We conducted experiments of different downstream tasks containing linear evaluation, fine-tuning, and transfer learning.
\subsection{Linear evaluation results with different small batch sizes}
In this subsection, we evaluate the performance of TriBYOL by performing linear evaluation on CIFAR-10, CIFAR-100, and STL-10 with different small batch sizes (32, 64, and 128).
The training epochs of self-supervised learning are set to 80.
Following the linear evaluation protocol from~\cite{chen2020simple}, we trained a linear classifier on top of the frozen ResNet50 backbone pretrained with TriBYOL and other comparative methods for 200 epochs and test for every 10 epochs.
Since self-supervised learning methods with negative sample pairs performed poorly even with a batch size of 128, we did not conduct experiments with a batch size of 32 or 64. 
\par
Table~\ref{tab1} shows the linear evaluation results with different small batch sizes.
``b32'', ``b64'', and ``b128'' denote the batch size of 32, 64, and 128, respectively.
From Table~\ref{tab1}, we can see that TriBYOL drastically outperforms all comparative methods in small-batch cases. 
Especially, the performance of TriBYOL even exceeds that of the pretrained model on ImageNet on CIFAR-10 and CIFAR-100 with small batch sizes.
When the batch size is 32, the training of SimSiam~\cite{chen2021exploring} collapses ($i.e.$, output was constant), and the test results are always a certain category, resulting in 1.00\% and 10.00\% on CIFAR-100 and STL-10, respectively.
Linear evaluation results on CIFAR-10, CIFAR-100, and STL-10 show that TriBYOL can learn better representations with different small batch sizes than other state-of-the-art self-supervised learning methods.
\subsection{Fine-tuning results with different numbers of labels}
In this subsection, we evaluate the performance of TriBYOL by fine-tuning a linear classifier with different numbers of labels on CIFAR-10, CIFAR-100, and STL-10.
The number of training epochs of self-supervised learning was set to 80.
We used the ResNet50 backbone that was trained with a batch size of 128.
We fine-tuned the linear classifier for 10 epochs and tested it for every epoch.
\par
Table~\ref{tab2} shows the fine-tuning results with different numbers of labels.
``1\%'', ``10\%'', and ``100\%'' denote the percentage of used labels.
We can see that TriBYOL outperforms all comparative methods with different numbers of labels.
Compared with training from scratch, TriBYOL shows better representation learning ability.  
Fine-tuning results on CIFAR-10, CIFAR-100, and STL-10 show that TriBYOL can learn better representations with different numbers of labels than other state-of-the-art self-supervised learning methods.
\subsection{Transfer learning results with different datasets}
In this subsection, we evaluate the performance of TriBYOL by training a linear classifier on top of the frozen representations learned from Tiny ImageNet on a variety of datasets containing MNIST, FashionMNIST, KMNIST, USPS, SVHN, CIFAR-10, CIFAR-100, and STL-10.
The number of training epochs of self-supervised learning was set to 80.
We use the ResNet50 backbone that was trained with a batch size of 128.
We trained the linear classifier for 200 epochs and tested it for every 10 epochs.
\par
Table~\ref{tab3} shows the transfer learning results on different datasets.
As shown in Table~\ref{tab3}, TriBYOL outperforms all comparative methods on different datasets. 
Furthermore, when transferring to MNIST, FashionMNIST, KMNIST, and USPS, the performance of TriBYOL even exceeds that of the pretrained model on ImageNet.
Transfer learning results on different datasets show that TriBYOL can learn more generalizable representations than other state-of-the-art self-supervised learning methods.
\section{Conclusion}
This paper has proposed a novel self-supervised learning method called TriBYOL.
The proposed method presents a new triplet network combined with a triple-view loss based on the state-of-the-art method BYOL for better representation learning performance with small batch sizes.
We conduct extensive experiments of different downstream tasks containing linear evaluation, fine-tuning, and transfer learning to verify the effectiveness of TriBYOL.
Experimental results show that our method can drastically outperform state-of-the-art self-supervised learning methods on several datasets in small-batch cases.
Although we only take experiments in small-batch cases due to the limitations of computing resources, we believe that our method can also have good performance even with large batch sizes.
Evaluation of TriBYOL on other high-resolution image datasets in the real world will be one of our future works.
\bibliographystyle{IEEEbib}
\bibliography{refs}

\begin{thebibliography}{10}

\bibitem{lecun2015deep}
Yann LeCun, Yoshua Bengio, and Geoffrey Hinton,
\newblock ``Deep learning,''
\newblock {\em Nature}, vol. 521, pp. 436--444, 2015.

\bibitem{liu2020self}
Xiao Liu, Fanjin Zhang, Zhenyu Hou, Zhaoyu Wang, Li~Mian, Jing Zhang, and Jie
  Tang,
\newblock ``Self-supervised learning: Generative or contrastive,''
\newblock {\em arXiv preprint arXiv:2006.08218}, 2020.

\bibitem{jing2020self}
Longlong Jing and Yingli Tian,
\newblock ``Self-supervised visual feature learning with deep neural networks:
  A survey,''
\newblock {\em IEEE Transactions on Pattern Analysis and Machine Intelligence},
  2020.

\bibitem{gidaris2018unsupervised}
Spyros Gidaris, Praveer Singh, and Nikos Komodakis,
\newblock ``Unsupervised representation learning by predicting image
  rotations,''
\newblock in {\em Proceedings of the International Conference on Learning
  Representations (ICLR)}, 2018.

\bibitem{noroozi2016unsupervised}
Mehdi Noroozi and Paolo Favaro,
\newblock ``Unsupervised learning of visual representations by solving jigsaw
  puzzles,''
\newblock in {\em Proceedings of the European Conference on Computer Vision
  (ECCV)}, 2016, pp. 69--84.

\bibitem{misra2020self}
Ishan Misra and Laurens van~der Maaten,
\newblock ``Self-supervised learning of pretext-invariant representations,''
\newblock in {\em Proceedings of the IEEE/CVF Conference on Computer Vision and
  Pattern Recognition (CVPR)}, 2020, pp. 6707--6717.

\bibitem{Bromley1993signature}
Jane Bromley, Isabelle Guyon, Yann LeCun, Eduard S\"{a}ckinger, and Roopak
  Shah,
\newblock ``Signature verification using a "siamese" time delay neural
  network,''
\newblock in {\em Proceedings of the Advances in Neural Information Processing
  Systems (NeurIPS)}, 1993.

\bibitem{tian2020understanding}
Yuandong Tian, Lantao Yu, Xinlei Chen, and Surya Ganguli,
\newblock ``Understanding self-supervised learning with dual deep networks,''
\newblock {\em arXiv preprint arXiv:2010.00578}, 2020.

\bibitem{li2021cross}
Guang Li, Ren Togo, Takahiro Ogawa, and Miki Haseyama,
\newblock ``Cross-view self-supervised learning via momentum statistics in
  batch normalization,''
\newblock in {\em Proceedings of the IEEE International Conference on Consumer
  Electronics – Taiwan (ICCE-TW)}, 2021.

\bibitem{tian2021understanding}
Yuandong Tian, Xinlei Chen, and Surya Ganguli,
\newblock ``Understanding self-supervised learning dynamics without contrastive
  pairs,''
\newblock in {\em Proceedings of the International Conference on Machine
  Learning (ICML)}, 2021.

\bibitem{chen2020simple}
Ting Chen, Simon Kornblith, Mohammad Norouzi, and Geoffrey Hinton,
\newblock ``A simple framework for contrastive learning of visual
  representations,''
\newblock in {\em Proceedings of the International Conference on Machine
  Learning (ICML)}, 2020.

\bibitem{chen2021exploring}
Xinlei Chen and Kaiming He,
\newblock ``Exploring simple siamese representation learning,''
\newblock in {\em Proceedings of the IEEE/CVF Conference on Computer Vision and
  Pattern Recognition (CVPR)}, 2021.

\bibitem{grill2020bootstrap}
Jean-Bastien Grill, Florian Strub, Florent Altch{\'e}, Corentin Tallec, Pierre
  Richemond, Elena Buchatskaya, Carl Doersch, Bernardo Avila~Pires, Zhaohan
  Guo, Mohammad Gheshlaghi~Azar, et~al.,
\newblock ``Bootstrap your own latent-a new approach to self-supervised
  learning,''
\newblock in {\em Proceedings of the Advances in Neural Information Processing
  Systems (NeurIPS)}, 2020, pp. 21271--21284.

\bibitem{li2021self}
Guang Li, Ren Togo, Takahiro Ogawa, and Miki Haseyama,
\newblock ``Self-supervised learning for gastritis detection with gastric x-ray
  images,''
\newblock {\em arXiv preprint arXiv:2104.02864}, 2021.

\bibitem{li2021triplet}
Guang Li, Ren Togo, Takahiro Ogawa, and Miki Haseyama,
\newblock ``Triplet self-supervised learning for gastritis detection with
  scarce annotations,''
\newblock in {\em Proceedings of the IEEE Global Conference on Consumer
  Electronics (GCCE)}, 2021.

\bibitem{li2022self2}
Guang Li, Ren Togo, Takahiro Ogawa, and Miki Haseyama,
\newblock ``Self-knowledge distillation based self-supervised learning for
  covid-19 detection from chest x-ray images,''
\newblock in {\em Proceedings of the IEEE International Conference on
  Acoustics, Speech and Signal Processing (ICASSP)}, 2022.

\bibitem{tao2020remote}
Chao Tao, Ji~Qi, Weipeng Lu, Hao Wang, and Haifeng Li,
\newblock ``Remote sensing image scene classification with self-supervised
  paradigm under limited labeled samples,''
\newblock {\em IEEE Geoscience and Remote Sensing Letters}, 2020.

\bibitem{tian2020contrastive}
Yonglong Tian, Dilip Krishnan, and Phillip Isola,
\newblock ``Contrastive multiview coding,''
\newblock in {\em Proceedings of the European Conference on Computer Vision
  (ECCV)}, 2020.

\bibitem{hoffer2015deep}
Elad Hoffer and Nir Ailon,
\newblock ``Deep metric learning using triplet network,''
\newblock in {\em Proceedings of the International Workshop on Similarity-based
  Pattern Recognition}, 2015, pp. 84--92.

\bibitem{antti2017mean}
Antti Tarvainen and Harri Valpola,
\newblock ``Mean teachers are better role models: Weight-averaged consistency
  targets improve semi-supervised deep learning results,''
\newblock in {\em Proceedings of the Advances in Neural Information Processing
  Systems (NeurIPS)}, 2017, pp. 1195--1204.

\bibitem{he2016deep}
Kaiming He, Xiangyu Zhang, Shaoqing Ren, and Jian Sun,
\newblock ``Deep residual learning for image recognition,''
\newblock in {\em Proceedings of the IEEE/CVF Conference on Computer Vision and
  Pattern Recognition (CVPR)}, 2016, pp. 770--778.

\bibitem{lecun2010mnist}
Yann LeCun, Corinna Cortes, and CJ~Burges,
\newblock ``Mnist handwritten digit database,''
\newblock 2010.

\bibitem{xiao2017fashion}
Han Xiao, Kashif Rasul, and Roland Vollgraf,
\newblock ``Fashion-mnist: a novel image dataset for benchmarking machine
  learning algorithms,''
\newblock {\em arXiv preprint arXiv:1708.07747}, 2017.

\bibitem{clanuwat2018deep}
Tarin Clanuwat, Mikel Bober-Irizar, Asanobu Kitamoto, Alex Lamb, Kazuaki
  Yamamoto, and David Ha,
\newblock ``Deep learning for classical japanese literature,''
\newblock {\em arXiv preprint arXiv:1812.01718}, 2018.

\bibitem{hull1994database}
Jonathan~J. Hull,
\newblock ``A database for handwritten text recognition research,''
\newblock {\em IEEE Transactions on Pattern Analysis and Machine Intelligence},
  vol. 16, no. 5, pp. 550--554, 1994.

\bibitem{netzer2011reading}
Yuval Netzer, Tao Wang, Adam Coates, Alessandro Bissacco, Bo~Wu, and Andrew~Y
  Ng,
\newblock ``Reading digits in natural images with unsupervised feature
  learning,''
\newblock in {\em Proceedings of the Advances in Neural Information Processing
  Systems (NeurIPS), Workshop}, 2011.

\bibitem{krizhevsky2009learning}
Alex Krizhevsky, Geoffrey Hinton, et~al.,
\newblock ``Learning multiple layers of features from tiny images,''
\newblock 2009.

\bibitem{coates2011analysis}
Adam Coates, Andrew Ng, and Honglak Lee,
\newblock ``An analysis of single-layer networks in unsupervised feature
  learning,''
\newblock in {\em Proceedings of the International Conference on Artificial
  Intelligence and Statistics (AISTATS)}, 2011, pp. 215--223.

\bibitem{le2015tiny}
Ya~Le and Xuan Yang,
\newblock ``Tiny imagenet visual recognition challenge,''
\newblock 2015.

\bibitem{deng2009imagenet}
Jia Deng, Wei Dong, Richard Socher, Li-Jia Li, Kai Li, and Li~Fei-Fei,
\newblock ``Imagenet: A large-scale hierarchical image database,''
\newblock in {\em Proceedings of the IEEE Conference on Computer Vision and
  Pattern Recognition (CVPR)}, 2009, pp. 248--255.

\end{thebibliography}

\end{document}